\documentclass[letterpaper]{article} 
\usepackage{aaai2026}  
\usepackage{times}  
\usepackage{helvet}  
\usepackage{courier}  
\usepackage[hyphens]{url}  
\usepackage{graphicx} 
\usepackage{natbib}  
\usepackage{caption} 
\usepackage{algorithm}
\usepackage{algorithmic}

\usepackage{newfloat}
\usepackage{listings}
\DeclareCaptionStyle{ruled}{labelfont=normalfont,labelsep=colon,strut=off} 
\floatstyle{ruled}
\newfloat{listing}{tb}{lst}{}
\floatname{listing}{Listing}
\usepackage{booktabs}
\usepackage{multirow}
\usepackage[table]{xcolor}
\usepackage{amsmath}
\usepackage{tikz}
\usetikzlibrary{arrows.meta}
\usepackage[most]{tcolorbox}
\tcbset{
  colback=gray!10,
  colframe=black!60,    
  boxrule=0.8pt,        
  arc=4pt,            
  left=6pt, right=6pt, top=6pt, bottom=6pt,
  enhanced jigsaw,
}

\usepackage{pdfpages}
\usepackage{placeins}

\title{SlideBot: A Multi-Agent Framework for Generating Informative, Reliable, Multi-Modal Presentations}
\author {
    Eric Xie\textsuperscript{\rm 1},
    Danielle Waterfield\textsuperscript{\rm 1},
    Michael Kennedy\textsuperscript{\rm 1},
    Aidong Zhang\textsuperscript{\rm 1}
}
\affiliations {
    \textsuperscript{\rm 1}University of Virginia\\
    \{jrg4wx, dla9ck, mjk3p, aidong\}@virginia.edu
}

\begin{document}

\maketitle

\begin{abstract}

Large Language Models (LLMs) have shown immense potential in education, automating tasks like quiz generation and content summarization. However, generating effective presentation slides introduces unique challenges due to the complexity of multimodal content creation and the need for precise, domain-specific information. Existing LLM-based solutions often fail to produce reliable and informative outputs, limiting their educational value. To address these limitations, we introduce SlideBot - a modular, multi-agent slide generation framework that integrates LLMs with retrieval, structured planning, and code generation. SlideBot is organized around three pillars: \textbf{informativeness}, ensuring deep and contextually grounded content; \textbf{reliability}, achieved by incorporating external sources through retrieval; and \textbf{practicality}, which enables customization and iterative feedback through instructor collaboration. It incorporates evidence-based instructional design principles from Cognitive Load Theory (CLT) and the Cognitive Theory of Multimedia Learning (CTML), using structured planning to manage intrinsic load and consistent visual macros to reduce extraneous load and enhance dual-channel learning. Within the system, specialized agents collaboratively retrieve information, summarize content, generate figures, and format slides using \LaTeX, aligning outputs with instructor preferences through interactive refinement. Evaluations from domain experts and students in AI and biomedical education show that SlideBot consistently enhances conceptual accuracy, clarity, and instructional value. These findings demonstrate SlideBot's potential to streamline slide preparation while ensuring accuracy, relevance, and adaptability in higher education.

\end{abstract}

\textbf{Extended Version} --- link to ArXiv paper 

\section{Introduction}

Artificial intelligence (AI) has transformed education by automating tasks such as quiz generation and content summarization, reducing instructor workloads, and improving student learning experiences \citep{elkins2024how, agrawal2024cyberq, fagbohun2024beyond}. Among these advancements, large language models (LLMs) stand out due to their ability to follow user instructions \citep{ouyang2022training} and effectively process contextual information, allowing them to perform complex tasks in diverse domains \citep{achiam2023gpt,touvron2023llama2}. Their proficiency in understanding and generating text unlocks significant potential for automating tasks in education.

Slide presentations are a core instructional medium that facilitates structured and engaging multimedia content that supports deeper student learning \citep{alley2005rethinking, bartsch2003effectiveness, mayer2005introduction}. However, generating slides is time-consuming and demands coordination across multiple modalities, text, visuals, and layout, a task that challenges existing LLM-based solutions. Without access to external information, LLMs rely on their parametric knowledge, the internal representations learned from their training data. In other words, the models draw on the learned statistical patterns and associations to generate content. This increases the risk of generating outdated or hallucinated content, statements that seem credible but lack factual accuracy, raising reliability concerns in educational contexts \citep{flier2023publishing,cremin2022big, bender2021dangers}. 
These limitations are especially problematic in evolving fields such as biomedicine or AI, where domain-specific knowledge is complex and rapidly updated \citep{ahmad2023creating}.

In addition to using factual domain-specific knowledge, pedagogical knowledge is key when generating slides, as effective instructional materials must align with how people learn and how teaching practices are mediated by tools and contexts. The importance of pedagogical knowledge is evidenced by cognitive load theory (CLT) and the cognitive theory of multimedia learning (CTML) \citep{chandler1991cognitive, mayer2005cognitive}. CLT posits that one’s working memory has an overall limited capacity for processing new information, which is divided into three types: intrinsic (complexity of new information), extraneous (distractions from new information), and germane (effort used to process and integrate new information) \citep{chandler1991cognitive}. Effectively designed slides optimize overall cognitive load by reducing or eliminating distractions to support germane load and promote schema construction \citep{paas20142}. Automating slide generation with LLMs helps reduce extraneous load by embedding evidence-based design principles such as eliminating redundant text and aligning visuals with key points that ultimately minimize unnecessary processing demands. While CLT emphasizes managing overall cognitive resources, CTML extends this by focusing on how learners process information across channels as well as how well-designed multimedia materials can optimize this processing to support meaningful learning \citep{mayer2002multimedia, mayer2005introduction}. CTML posits that learners process information through dual visual and auditory channels with limited working memory and meaningful learning occurs when materials support the selection, organization, and integration of information \citep{mayer2002multimedia,mayer2005introduction}. Poorly designed slides can allow cognitive overload to occur. As such, automating slide creation with LLMs offers a way to embed CTML principles such as coherence (removing unnecessary information), signaling (highlighting important information), and spatial contiguity (aligning text and visuals), directly into the design process while aiming to avoid cognitive overload.

To address challenges faced by educators in generating slides, we present SlideBot: an agentic slide generation framework that embeds the theoretical aspects of CLT and CTML and leverages large language models in combination with retrieval, structured planning, and modular code generation. SlideBot decomposes the generation process across multiple specialized agents - including retrievers, planners, figure creators, and coders - to produce well-formatted slides from retrieved content and pedagogical prompts. The slides are made using  \LaTeX{}\footnote{\url{https://www.latex-project.org}}-based slides using the Beamer\footnote{\url{https://ctan.org/pkg/beamer}} package, combining precise formatting with flexible customization. This structured architecture enables the consistent generation of context-grounded, university-level presentations tailored for instructional use.

We design SlideBot around three core pillars: \textbf{informativeness}, ensuring deep, domain-specific coverage; \textbf{reliability}, grounding outputs in high-quality external sources; and \textbf{practicality}, supporting usability and instructor customization. We validate our approach through both student surveys and expert reviews, demonstrating consistent gains in explanation quality, conceptual accuracy, and overall suitability over Microsoft Copilot, a state-of-the-art presentation generation tool, as well as a direct prompt baseline, where the model produces slides directly from a single prompt without retrieval, planning, or other agentic assistance. By incorporating credible sources and structuring outputs around best practices in multimedia design, our system upholds the pillars by mitigating hallucinations, improving explanation clarity, and providing an adaptable pipeline that instructors can customize to their content, formatting, and pedagogical needs. This interactive, modular approach establishes a new paradigm for reliable and instructor-friendly AI-assisted presentation generation.

\section{Related Work}

\subsection{Artificial Intelligence in Education}

The integration of artificial intelligence (AI) into education has revolutionized modern classrooms, offering tools that enhance both teaching and learning experiences. AI-powered systems in education can be broadly categorized based on their intended end user: student-centered AI, which focuses on improving student learning, and educator-centered AI, which streamlines teaching workflows through automation \citep{wang2024large}.

\paragraph{Student-Centered Learning Support.}
Language models have been shown to provide detailed explanations and adapt feedback to individual learning needs. For instance, GPT-based systems can rival or exceed the explanatory capabilities of students or teaching assistants in specific domains. \citet{balse2023evaluating} found that GPT-3.5's explanations for programming errors matched the quality of TA-generated feedback. Similarly, \citet{leinonen2023comparing} showed that GPT-3 produced clearer code explanations than students. However, other studies reveal limitations in factual accuracy and pedagogical depth: \citet{prihar2023comparing} observed that GPT-3 explanations for middle school math problems fell short of those written by teachers, reinforcing concerns about perpetuating misconceptions  \citep{kunz2024properties}. To improve interactivity, recent systems adopt dialogic strategies such as Socratic prompting and recontextualization to align explanations with student interests \citep{shridhar2022automatic, yadav2023contextualizing}. AI-powered chatbots like EduChat \citep{dan2023educhat} and the Taiwan Adaptive Learning Platform \citep{kuo2023leveraging} integrate real-time feedback and dynamic questioning, enhancing student engagement through continual adaptation.

\paragraph{Educator-Centered Support and Content Generation.}
Educator-facing tools have focused on reducing instructional overhead by automating tasks such as quiz creation, grading, and feedback generation \citep{alsafari2024effective, kasneci2023chatgpt}. For example, CyberQ \citep{agrawal2024cyberq} uses LLMs with knowledge graph augmentation to generate targeted cybersecurity assessments. Other systems demonstrate that LLM-generated quiz questions can rival those written by instructors in clarity and pedagogical quality \citep{elkins2024how, doughty2024comparative, xiao2023evaluating}. Automated feedback systems further show that LLMs can produce high-quality evaluative comments across disciplines \citep{xiao2024automation, li2024automated}.

\subsection{Automated Slide Development}
Automating slide generation has gained significant interest, particularly for scientific and technical presentations. Early work in this area focused on extractive methods that identify and rank important sentences from documents to form slide content. For example, \citet{sefid2019automatic} proposed a method based on the SummaRuNNer model \citep{nallapati2017summarunner}, which uses a windowed labeling ranking system to combine semantic and lexical features within a sentence window, measuring the importance and novelty of sentences for slide construction.

Other researchers explored alternative approaches to identify relevant information. \citet{hu2015ppsg} developed PPSGen, a framework that uses Support Vector Regressors and Integer Linear Programming (ILP) to rank and select key sentences. \citet{wang2017phrase} proposed a phrase-based approach that extracts key phrases and learns hierarchical relationships to structure bullet points for slides. Over time, these strategies evolved to incorporate deep learning methods. For example, \citet{sefid2021extractive} extended their earlier work by incorporating contextual information and deep neural network approaches to enhance sentence scoring and summary construction. Their approach combines feature-based and neural methods to improve coherence and relevance.

Modern advancements in this field focus on interactive and real-time solutions using up-to-date information. For example, to address the challenge of handling longer documents, \citet{gupta2023automatic} investigated the use of LLMs with extended token limits, such as Longformer-Encoder-Decoder (LED) \citep{beltagy2020longformer} and BIGBIRD-Pegasus \citep{zaheer2020big}. These models process full-length scientific papers and generate cohesive section-slide pairs. Microsoft Copilot \citep{microsoft365copilot2023} integrates the LLM capabilities directly into PowerPoint to assist users with slide creation by generating slide outlines, suggested text, and visual content based on user prompts and document context. This approach highlights a trend toward more flexible user-driven slide generation that automates many aspects of presentation design.

\begin{figure*}
    \centering
    \includegraphics[width=0.9\textwidth]{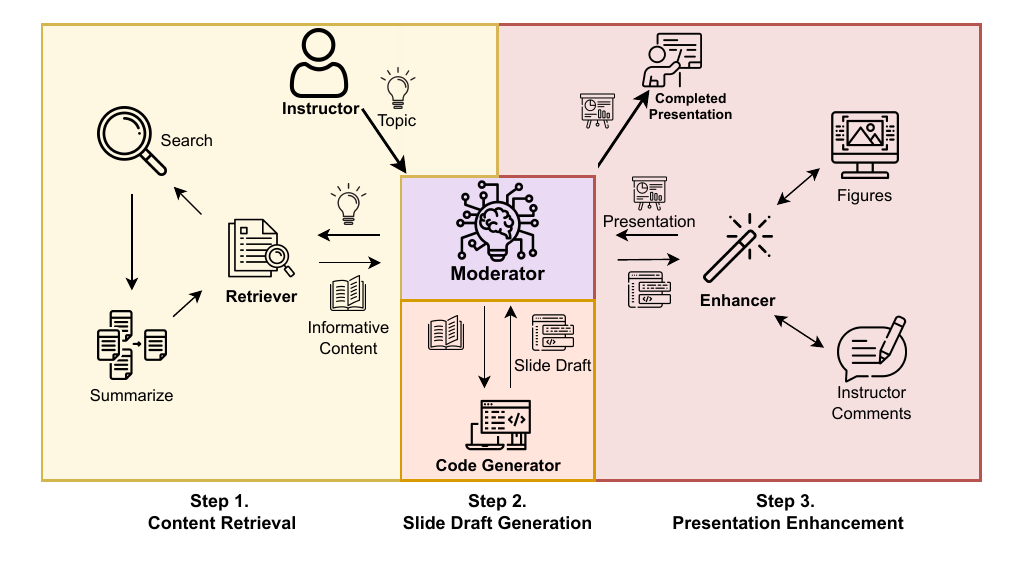}
    \caption{SlideBot's slide generation pipeline operates in three stages: Content Retrieval, where the Moderator receives a topic from the Instructor and communicates with the Retriever to gather and summarize relevant information from a user-selected or automatically designated corpus; Slide Draft Generation, where the Moderator constructs a structured slide plan and the Code Generator translates it into \LaTeX  Beamer code; and Presentation Enhancement, where the Enhancer inserts figures and instructional comments before returning a completed presentation to the instructor. The Moderator coordinates all agents and manages feedback loops to ensure quality, adaptability, and consistency.}
    
    \label{fig:pipeline}
\end{figure*}

\subsection{Agentic and Instructor-Aligned Generation Frameworks}

Recent advances in AI have shifted from single-prompt generation to structured, multi-agent pipelines in which distinct components collaborate to solve complex tasks. This agentic paradigm enables systems to decompose tasks such as code generation \citep{yang2024swe}, tool use \citep{schick2023toolformer}, and long-context reasoning \citep{lu2023chameleon} into subtasks handled by specialized agents. In education, such modularity can allow for a closer alignment with instructional workflows, supporting features like structured planning, instructor comments, and multimodal content insertion.

Among these agents, retrieval modules often leverage techniques from Retrieval-Augmented Generation (RAG) \citep{lewis2020retrieval}, which grounds model responses in external documents to improve factual accuracy and reduce hallucinations. RAG has proven effective in domains requiring up-to-date, trustworthy information, including biomedical QA \citep{jin2023retrieve, lala2023paperqa} and literature synthesis \citep{jiang2023active, mialon2023augmented}.

Our proposed SlideBot builds on this paradigm by coordinating a set of specialized components under a central Moderator to handle retrieval, structured slide planning, and presentation enhancement. This modular design enables iterative refinement, ensuring that outputs remain grounded in domain knowledge while following evidence-based instructional principles to support learning.

\section{Methodology} \label{sec:method}

SlideBot is built around three key pillars to support effective university-level content creation:

\begin{itemize}
    \item \textbf{Informativeness}: Combines delivery clarity, factual accuracy, and breadth of domain-relevant subtopics to provide a coherent narrative that supports student learning.
    \item \textbf{Reliability}: Presents factually-accurate content grounded in credible sources, maintaining consistent results across each generated presentation. 
    \item \textbf{Practicality}: Readily usable and adaptable in real teaching environments, providing clear formatting, multimodal support, and instructor-facing features. 
\end{itemize}

To fulfill these pillars, SlideBot employs a multimodal, multi-agent framework that decomposes educational slide generation into three stages: Content Retrieval, Slide Draft Generation, and Presentation Enhancement. Each stage is handled by specialized agents coordinated by a central Moderator, as shown in Figure~\ref{fig:pipeline}.

\subsection{Content Retrieval}

The Content Retrieval stage is responsible for gathering relevant, domain-specific information to ground the presentation in accurate and informative content using Retrieval-Augmented Generation (RAG) \citep{lewis2020retrieval}. RAG combines generative capabilities of LLMs with external retrieval mechanisms to reduce hallucinations and enhance factual grounding \citep{bechard2024reducing}. Our system implements a modular corpus interface, allowing retrieval from interchangeable knowledge sources, such as research papers or textbooks. This ensures adaptability across disciplines, supporting both practicality and reliability.

Once a target corpus is selected, the Retriever agent constructs a query to the appropriate source. For academic literature, we use the arXiv API\footnote{\url{https://info.arxiv.org/help/api/index.html}} to identify relevant scholarly articles using keyword matching. For structured resources such as textbooks, we leverage BM-25 \citep{robertson2009probabilistic}, a probabilistic ranking algorithm that prioritizes documents where uncommon query terms appear frequently. This allows instructors to incorporate cutting-edge research and curated textbook materials into their presentations.

After retrieving relevant documents or passages, the Retriever constructs a detailed summary and finds source metadata for citation (e.g., title, authors, publication date). These summaries are passed back to the Moderator, which may select a subset of the content for inclusion into the slides. This filtering step ensures that only high-quality, relevant material is discussed within the presentation.

\subsection{Slide Draft Generation}

Once the Moderator receives the summaries from the retrieval stage, it constructs a detailed slide plan to guide the generation of content. This plan follows a predefined structural guide that provides a flexible outline while embedding evidence-based instructional principles from cognitive load theory (CLT) and the cognitive theory of multimedia learning (CTML). The guide begins with a high-level introduction to manage intrinsic load by establishing core concepts, progresses into deeper exploration organized by either individual papers or conceptual subtopics to support schema construction, and concludes with key takeaways that reinforce germane load. Additional structural elements such as a title slide, table of contents, and references are also included to ensure presentation flow and reduce extraneous load. The slide plan also incorporates CTML principles: signaling is implemented by using bolded key terms and bullet hierarchies to direct attention to main ideas, coherence is maintained by limiting each slide to concise, directly relevant points with supporting visuals, and spatial contiguity is achieved by pairing explanatory text with corresponding diagrams. Using this structural guide, the Moderator constructs a slide plan that specifies individual slide headers, key facts or explanations to present on each slide, and corresponding citations.

After the slide plan is completed, it is passed to the Code Generator agent. Due to the complexity of direct multimodal generation, where language models often struggle to simultaneously produce coherent text, layout, and visual structure, we instead leverage modern LLMs' strengths in code generation \citep{jiang2024survey}. The Code Generator translates the structured blueprint into \LaTeX  code that creates structured presentations when compiled. \LaTeX  is widely adopted within academia for its precision, flexibility, and broad collection of packages. We use the Beamer class, a \LaTeX  package specialized for slide presentations to structure the output, ensuring consistent formatting, support for mathematical notation, citations, and the integration of dynamic visual elements. After the code is compiled and validated by the Moderator, the code is returned to the Code Generator if a compilation error occurs, along with the error message and suggested revisions. This iterative loop minimizes hallucinations in code generation and ensures reliable, executable slide output.

\begin{figure*}
    \centering
    \includegraphics[width=\linewidth]{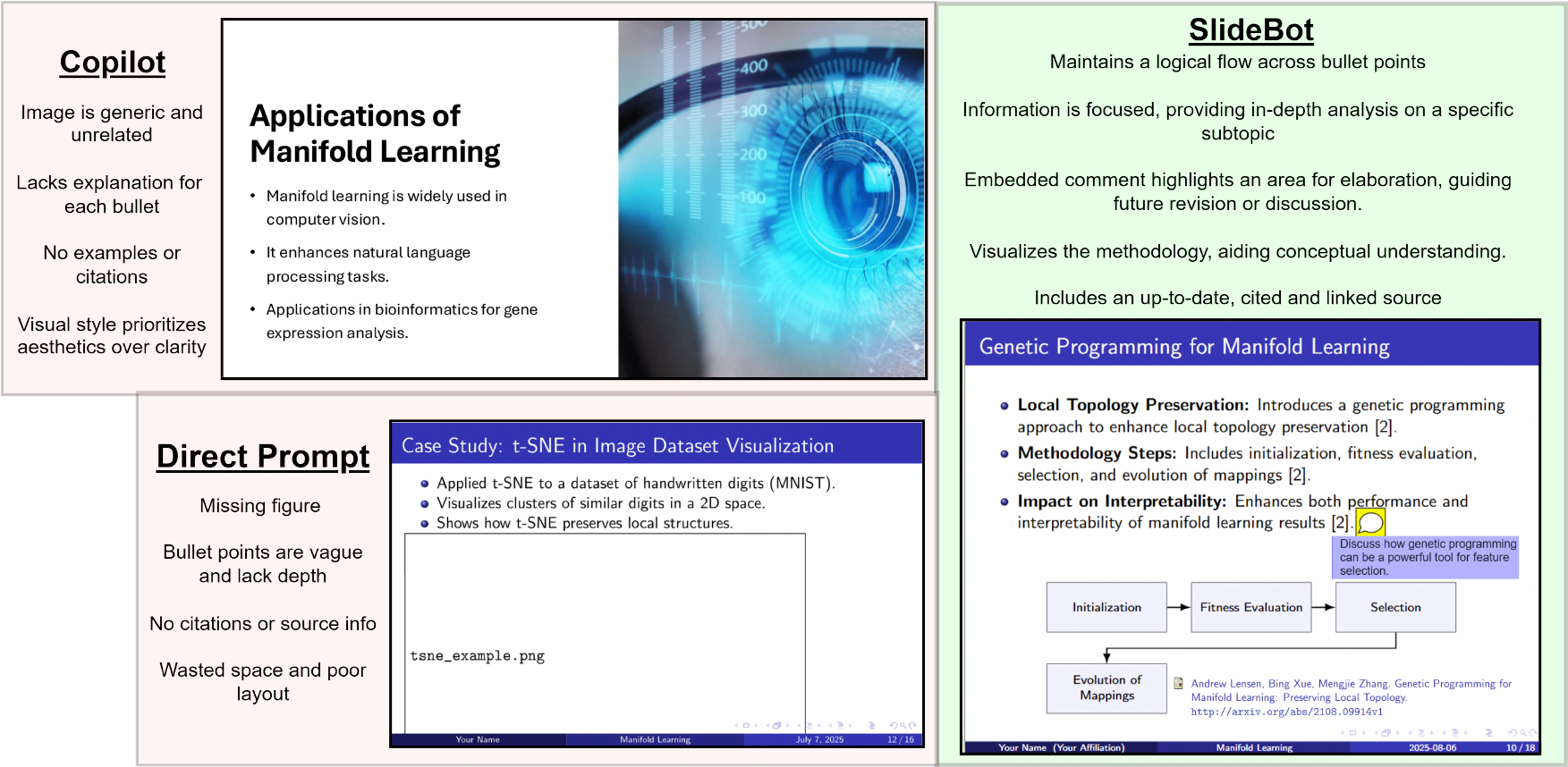}
    \caption{Qualitative comparison of slides generated by Copilot and GPT-4o Direct Prompting (left), and SlideBot (right) on the topic “Manifold Learning.” Copilot and Direct Prompt outputs lack explanatory depth, meaningful visuals, and relevant citations. In contrast, SlideBot produces focused, grounded, and pedagogically useful content by retrieving information from \citet{lensen2021genetic} via the Retriever agent, adding figures and instructor comments via the Enhancer, and compiling structured \LaTeX{} Beamer slides through the Code Generator.}i
    \label{fig:demo_figure}
\end{figure*}

\begin{table*}[h]
\centering
\begin{tabular}{llcc|c}
\toprule
\textbf{Pillar} & \textbf{Metric} & \textbf{Copilot} & \textbf{SlideBot (Ours)} & $\boldsymbol{\Delta}$ \\
\midrule
\textbf{Informativeness}
  & Explanation Style     & 2.24 & \textbf{3.96} & +1.71 \\
\rowcolor{gray!25}
  & Conceptual Accuracy   & 2.71 & \textbf{3.57} & +0.86 \\
\rowcolor{gray!25}
  & Topic Coverage             & 2.67 & \textbf{4.10} & +1.43 \\
\midrule
\textbf{Reliability} 
& Credibility & 1.93 & \textbf{4.36} & +2.42 \\
& Variability* & 1.47 & \textbf{0.33} & –1.14 \\

\midrule
\textbf{Practicality}
  & Structure \& Flow     & 3.42 & \textbf{4.04} & +0.62 \\
  & Overall Suitability   & 2.09 &\textbf{ 3.67} & +1.58 \\
\rowcolor{gray!25}
  & Instructor Utility    & 2.14 & \textbf{3.76} & +1.62 \\
\bottomrule
\end{tabular}
\caption{Comparison of SlideBot and Copilot across three evaluation pillars: \textit{Informativeness}, \textit{Reliability}, and \textit{Practicality}. Metrics are derived from both a student survey (unshaded rows) and an expert evaluation (shaded rows). All scores are averaged on a 1–5 scale. $\boldsymbol{\Delta}$ displays the difference between SlideBot's and Copilot's score for each metric. *The "Variability" metric captures the difference in average "Overall Suitability" between the best- and worst-rated presentations for each method - lower values are desirable, indicating more consistent outputs.}
\label{tab:main_eval}
\end{table*}

\subsection{Presentation Enhancement}

During the final stage of the pipeline, the Moderator identifies candidate slides where visual elements would enhance signaling or coherence, such as diagrams, charts, or architectural overviews, to be inserted by the Enhancer agent. To ensure consistency, the Enhancer uses prewritten figure macros, streamlining the process so that generating figures requires only supplying the appropriate parameters. The Enhancer also inserts instructor-view-only comments that provide instructional guidance, such as elaboration prompts, warnings about potential misconceptions, or suggestions for visual aids. These enhancements support both practicality, by aligning slides with classroom needs, and reliability, ensuring consistent figure generation. 

Once enhancement is complete, the finished slide deck is returned to the user. If desired, the user can send revision requests back to the Moderator, which coordinates with the appropriate agents to implement modifications. This iterative loop allows the slides to be refined multiple times, ensuring alignment with the instructor’s needs. 

SlideBot's modular design enables straightforward extensions, such as integrating additional specialized agents, expanding the library of figure macros, or incorporating new retrieval corpora to broaden domain coverage. Through its modular and iterative design, not only does this framework deliver accurate, pedagogically grounded content, SlideBot also remains adaptable, ready to incorporate new tools, agents, and resources as instructional needs evolve.

\section{Experiments}

\subsection{Experimental Settings}

To evaluate SlideBot's effectiveness, we conducted two studies in distinct domains: computer science and biomedical education. Each study employed the same dual-survey evaluation framework to assess presentation quality from both learner and instructor perspectives.

\paragraph{Survey Design and Evaluation Metrics.}
We designed two complementary surveys to capture feedback from both learners and instructors. Our general student survey targeted university students across disciplines to evaluate presentation quality from a learner’s perspective. It focused on surface-level aspects such as clarity, structure, and overall appeal, factors independent of domain knowledge. We surveyed 15 participants for the computer science student survey and 13 participants for the biomedicine student survey. Our expert survey was administered to domain-specific experts (graduate students and professors from the field) to reflect an instructor’s perspective, emphasizing the depth, accuracy, and instructional usefulness of the content. We surveyed 7 experts in computer science and 4 in biomedicine.

\paragraph{Computer Science Study.}
We curated a diverse set of undergraduate- to graduate-level AI-related topics, including manifold learning, attention mechanisms, and graph neural networks. To isolate the pipeline’s contributions from model capacity, we implemented our pipeline using GPT-4o-mini and compared it to GitHub Copilot, a state-of-the-art AI assistant linked directly to Microsoft PowerPoint.

\paragraph{Biomedical Study.}
We applied the pipeline to biomedical education by retrieving from a curated textbook corpus \citep{jin2021disease}, reflecting material used regularly in practice. We conducted two studies: (1) evaluating the impact of retrieved content on informativeness, and (2) varying model size to compare the effects of model scaling. In these experiments, we compare SlideBot to a Direct Prompt baseline, where the model receives a single instruction to “Generate a graduate-level presentation on [topic]” and produces slides without retrieval, planning, or specialized agents. Direct prompting reflects how AI models are typically used in practice, relying on the model’s general knowledge without further customization.

To assess system performance, we structured evaluation metrics around three core pillars:

\begin{itemize}
\item \textbf{Informativeness}: measured through explanation style (student), conceptual accuracy (expert), and topic coverage (expert), reflecting how clearly and accurately the material conveys a broad range of key ideas.
\item \textbf{Reliability}: assessed via credibility (student) and variability in suitability across topics (student), capturing trustworthiness and output consistency.
\item \textbf{Practicality}: evaluated through structure and flow (student), overall suitability (student), and instructor utility (expert), measuring real-world usability and alignment with teaching needs.
\end{itemize}

In each survey, participants reviewed three presentations per generation method and scored each metric on a scale of 1-5. This dual perspective evaluates both the quality and adaptability of generated presentations across multiple domains, model sizes, and pipeline configurations.

\subsection{Qualitative Analysis}

To ground these evaluations in a concrete example, Figure~\ref{fig:demo_figure} presents a comparison of slides generated by Copilot, GPT-4o-mini with a direct prompt (single instruction, no structured assistance), and our full pipeline (SlideBot). The Copilot output violates several evidence-based instructional design principles. For example, irrelevant visuals and absent citations hinder both the coherence principle (integrating only relevant information) and the signaling principle (highlighting key concepts). Vague bullet points impose unnecessary extraneous load under Cognitive Load Theory (CLT) by forcing learners to infer missing context, while poor spatial alignment between text and visuals violates the spatial contiguity principle, reducing integration between modalities. Generation from a direct prompt results in similar issues, though its failures extend beyond instructional design principles. Without agent-based coordination, functional errors begin to occur, such as missing images or improperly formatted elements that extend past slide boundaries.

In contrast, SlideBot’s output applies CTML and CLT principles more effectively. Focused bullet points with citations reduce extraneous load and support reliability. Visuals are directly tied to the described methodology, fulfilling the coherence and multimedia principles and aiding germane load through schema construction. Embedded instructor comments act as signaling devices, providing strategies to direct learner attention toward key areas for elaboration or deeper reasoning. The logical flow across slides aligns with segmenting and pre-training principles, easing the transition from introductory material to deeper exploration. These qualitative improvements reflect our three design pillars - informativeness, reliability, and practicality - and illustrate the qualitative improvements that underpin the quantitative gains reported in the following section.

\subsection{Comparison with Copilot}

Table~\ref{tab:main_eval} summarizes average scores across all metrics split into their respective pillars, drawn from both the student and expert surveys.

SlideBot outperforms Copilot in all metrics related to informativeness. Explanation style, rated by students, shows the largest improvement with a gain of +1.71. Conceptual accuracy also increases substantially (+0.86), which is particularly noteworthy in an educational setting, as it reflects not only the correctness of individual facts but also the alignment of presented material with accepted domain knowledge and learning objectives. Inaccuracies or misconceptions from hallucinations can significantly hinder understanding and lead to persistent misunderstandings, an issue SlideBot helps address. SlideBot also scores higher in Topic Coverage, reflecting the inclusion of a broader range of subtopics. While coverage alone is often a matter of preference, in combination with strong conceptual accuracy and related measures, it indicates that SlideBot can effectively address a wide scope of material.

For reliability, the credibility of SlideBot's slides was rated +2.42 higher than Copilot’s. Additionally, our system achieved a lower variability score (–1.14), suggesting more consistent presentation quality across different topics. This is attributed to SlideBot's structured control over content planning, grounding in verifiable sources, and code validation loop, all of which help mitigate hallucinations and preserve factual consistency.

In terms of practicality, slides produced by our system were consistently viewed as more usable in real-world teaching settings. Improvements were observed in structure and flow (+0.62), overall suitability (+1.58), and instructor utility (+1.62). Instructor comments, consistent formatting, and figure support contributed to these gains, helping instructors better understand, adapt, and present the material.

Taken together, these results demonstrate that our pipeline substantially improves presentation quality across all dimensions, highlighting the value of a modular, retrieval-augmented, and planning-driven approach.

\begin{figure}[t]
\centering
\includegraphics[width=0.98\linewidth]{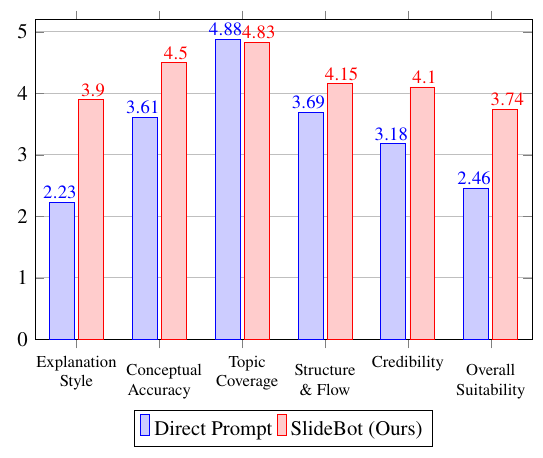}

\caption{Comparison of Direct Prompt and SlideBot generation (both using GPT-4o) across six presentation quality metrics, with Explanation Style, Structure \& Flow, Credibility, and Overall Suitability obtained from a student survey, and Conceptual Accuracy and Topic Coverage obtained from an expert survey. }
\label{fig:ablation_quality}
\end{figure}

\begin{figure}[h]
\centering
\includegraphics[width=\linewidth]{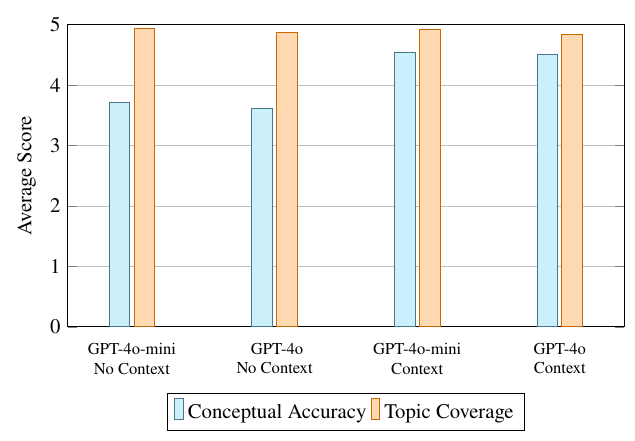}
\caption{Comparison of informativeness metrics across model and context combinations obtained from an expert survey. Results for GPT-4o models (with and without context) are repeated from Figure~\ref{fig:ablation_quality} to enable side-by-side comparison with GPT-4o-mini variants, renamed for clarity.}
\label{fig:ablation_modelsize}

\end{figure}

\subsection{Impact of Pipeline Guidance} 

To isolate the impact of our full architecture on \textit{Practicality} and \textit{Reliability}, we compare SlideBot's output to a Direct Prompt baseline slide generation that relies on the parametric capabilities of the language model. In this baseline, slides are generated in a single prompt without retrieval, structured planning, or enhancement. 

As shown in Figure \ref{fig:ablation_quality}, SlideBot significantly outperforms the Direct Prompt baseline across nearly all metrics. Explanation style improves by +1.67 points, while conceptual accuracy and credibility increase by +0.89 and +0.92, respectively. Direct Prompt achieves a similarly high Topic Coverage score, suggesting that the choice of base model may have a greater influence on topic breadth than the framework. The gap in Structure \& Flow is slightly narrower; however, without additional guidance, the Direct Prompt content still lacks utility as reflected in the +1.28 advantage for SlideBot in Overall Suitability. This contrast highlights that decomposing the task across specialized agents results in strong performance across all dimensions. 

\subsection{Impact of Model Size}

To assess whether our approach provides substantial benefits compared to model scaling, we compare GPT-4o and GPT-4o-mini with and without retrieved context, with data gathered from an expert survey. As shown in Figure \ref{fig:ablation_modelsize}, Topic Coverage remains consistent across all settings. Clear distinctions emerge in Conceptual Accuracy: GPT-4o-mini improves from 3.71 to 4.54 (+0.83), and GPT-4o from 3.61 to 4.50 (+0.89) when using a Retrieval agent. Interestingly, GPT-4o-mini marginally outperforms GPT-4o. We observe that because the smaller model has more limited parametric knowledge, it relies more heavily on the retrieved content and thus adheres more closely to it, whereas the larger model can default to its internal knowledge, which may be less explanatory or reliable.

Overall, the results reinforce that SlideBot's architectural design has a significantly larger impact than the size of the base model in driving conceptual understanding and content quality, indicating a cost-effective improvement measure.

\section{Conclusion}
While LLMs show promise in education, they often fall short in generating structured, reliable, and domain-specific presentation materials. To address this, we introduce SlideBot, a modular, agent-based slide generation framework with multimodal capabilites, that emphasizes informativeness, reliability, and practicality through the integration of Cognitive Load Theory and Cognitive Theory of Multimedia Learning principles. By combining retrieval, planning, and LaTeX-based formatting, SlideBot produces high-quality, customizable slides aligned with instructor needs in any subject area. Empirical results show significant improvements in informativeness, reliability, and practicality compared to Microsoft's Copilot, with benefits that stem from its adaptable, modular design rather than relying solely on larger models. As educational technology continues to evolve, our approach serves as a promising step toward creating impactful, reliable, AI-driven teaching tools.

\section*{Acknowledgements}
This work is supported in part by the US Department of Education under grant H327S240013.  Any opinions, findings, conclusions or recommendations expressed in this material are those of the
author(s) and do not necessarily reflect the views of the US Department of Education.


\newpage
\appendix

\section{Appendix}
\label{sec:appendix}

\subsection{Implementation Details}

While our experiments use a specific configuration of SlideBot, the framework is intentionally modular and can be implemented in a variety of ways depending on instructional goals, computational resources, and desired output characteristics. Each agent offers multiple points of variation - ranging from alternative retrieval sources and planning strategies to different prompting methods and output formats. These are not merely technical details; such choices can significantly influence the trade-offs between accuracy, breadth, and stylistic alignment. In this section, we describe the configuration used in our experiments, outline potential alternative designs for each component, and discuss the benefits and drawbacks of these choices to help guide future adaptations of the framework. 

The core behavior of each SlideBot agent in our implementation is achieved through prompting large language models. Prompting enables rapid prototyping, requires no additional model training, and allows for easy iteration on behavior through small adjustments to the input text. In our experiments, we designed domain-specific, role-oriented prompts for each agent type and iteratively refined them based on pilot outputs. These prompts are included later in this appendix to support replication.

However, prompting is only one of several possible approaches to implementing these agents. For example, fine-tuning a base model on curated examples of high-quality educational content could provide greater stylistic consistency and reduce variance in output quality. That said, fine-tuning comes with substantial costs - both financial (model training at scale) and logistical (acquiring and annotating domain-specific training data). In educational contexts, data collection can be especially challenging, as content must meet accuracy, alignment, and accessibility requirements while avoiding copyright or licensing issues. Instruction tuning or reinforcement learning from human feedback (RLHF) could also be used to align outputs with specific pedagogical goals, such as scaffolding complexity or prioritizing certain types of visual aids. These methods, however, require even larger quantities of domain-relevant data and a dedicated feedback pipeline, which may not be feasible in many instructional settings. 

Given these constraints, we selected prompting for its accessibility and adaptability. It avoids the steep financial and labor demands of large-scale model retraining while remaining easy to share, replicate, and adapt to new instructional contexts or content areas. This makes it particularly well-suited for research environments, small institutions, and educators who need high-quality materials without the infrastructure required for fine-tuning or RLHF.

\begin{figure}
\begin{tcolorbox}[colback=gray!10,colframe=black!90,title=Moderator: Keyword Generation Prompt Template.]

You are an expert tasked with planning for a presentation on \{topic\}. 
Your goal is to generate a list of keywords that will be used to search for relevant information in textbooks to be included in the presentation. 
Your response should only contain the list of keywords separated by commas.

\end{tcolorbox}
\caption{Prompt Template for the Moderator to select keywords based on the presentation topic for retrieval. }
\label{fig:mod_keyword_prompt}
\end{figure}

\begin{figure}
\begin{tcolorbox}[colback=gray!10,colframe=black!90,title=Retriever: Source Summarization Prompt Template.]

Title: \{Title\} \\
Abstract: \{Abstract\} \\
Paper: \{Paper\}

You are a scientific assistant summarizing research papers for presentation.

Give a concise summary of this paper, as if explaining to a graduate-level student who is new to the topic, followed by an in-depth analysis of the paper's contributions, methodology, and results. Emphasize any key formulas or algorithms.

\end{tcolorbox}
\caption{Prompt Template for the Retriever to summarize retrieved content arXiv paper content. }
\label{fig:ret_summary_prompt}
\end{figure}

\subsubsection{Retriever Agent.}

The retrieval pipeline proceeds in four steps. Given a user topic, the Moderator first expands it into a targeted search query, using the prompt shown in Figure \ref{fig:mod_keyword_prompt}. These keywords drive a branch-by-source retrieval step: (i) for research topics, we query the arXiv API (ranked by relevance or recency) and collect paper metadata (title, authors, abstract, PDF link, updated date); and (ii) for foundational topics, we query a biomedical textbook corpus with BM25 to return short, high-salience snippets. Because textbook snippets are already concise, we do not summarize them further. For arXiv results, the Retriever produces a structured summary using the paper-aware prompt in Figure \ref{fig:ret_summary_prompt}, capturing the problem, methods, contributions, and any key formulas or algorithms. Using the prompt in Figure \ref{fig:mod_select_prompt}, the Moderator ranks the candidate items and selects a small set of sources/snippets for slide planning. This pipeline yields curated, citation-ready context with minimal noise.

\begin{figure}
\begin{tcolorbox}[colback=gray!10,colframe=black!90,title=Moderator: Source Selection Prompt Template.]

Given the following summaries, select only the ones most relevant and informative for a graduate-level presentation on the topic \{topic\}. 
Return them in ranked order from most to least relevant.

\end{tcolorbox}
\caption{Prompt Template for the Moderator to select the most useful retrieved content from the retrieved context. }
\label{fig:mod_select_prompt}
\end{figure}

\begin{figure*}
\begin{tcolorbox}[colback=gray!10,colframe=blue!70,title=Retriever Agent Case Study ,fonttitle=\bfseries]

\textbf{Summary:} This paper explores a novel approach to manifold learning using genetic programming, focusing on preserving local topology in high-dimensional datasets. By evolving functional mappings from the original data space to a lower-dimensional embedding, the authors aim to enhance interpretability and performance compared to traditional manifold learning methods. Their results indicate that this new method not only improves upon previous genetic programming techniques but also outperforms several baseline manifold learning algorithms.

\textbf{Contributions:}
The primary contribution of this research is the introduction of a genetic programming framework specifically designed for manifold learning that emphasizes the preservation of local topology. This is significant because many existing manifold learning methods, while effective in reducing dimensionality, often compromise on interpretability and the retention of local structures within the data. The authors demonstrate that their approach can yield better performance metrics on various tasks, particularly those where local neighborhood relationships are crucial.

\textbf{Methodology:}

The authors propose a genetic programming algorithm that evolves mappings from the original high-dimensional space to a lower-dimensional embedding space. The key steps in their methodology include:

\begin{itemize}
    
    \item 1. \textbf{Initialization:} A population of random functions is generated, which can be represented as trees (common in genetic programming).
    \item 2. \textbf{Fitness Evaluation:} Each function is evaluated based on how well it preserves local topology. This is quantified using metrics such as the preservation of distances or neighborhood relationships in the embedding space.
    \item 3. \textbf{Selection and Evolution:} Functions that perform better in preserving local topology are selected for reproduction. Genetic operations such as crossover and mutation are applied to create new functions.    
    \item 4. \textbf{Iteration:} The process is repeated over several generations, gradually evolving better-performing mappings.
\end{itemize}

\textbf{Results:}

The authors compare their genetic programming approach against several baseline manifold learning methods, such as t-SNE, UMAP, and PCA. They report that their method consistently outperforms these techniques, particularly in tasks where local structure is essential. The results are quantified using standard performance metrics, which may include reconstruction error, clustering quality, or visualization clarity.        

The findings suggest that the evolved mappings not only provide a lower-dimensional representation of the data but also retain interpretability, allowing researchers to understand the relationships and structures within the data more effectively. This is a crucial advancement in the field of exploratory data analysis, where understanding the underlying data structure is often as important as the analysis itself.

In conclusion, this paper presents a significant advancement in manifold learning through the innovative application of genetic programming, with promising implications for both performance and interpretability in high-dimensional data analysis.

\end{tcolorbox}
\caption{Summary of the paper "Genetic Programming for Manifold Learning: Preserving Local Topology" \citep{lensen2021genetic} output by the Retriever Agent for the presentation "Manifold Learning". A slide extracted from the final presentation is depicted in Figure \ref{fig:demo_figure}.}
\label{fig:summary_output}
\end{figure*}

Our experiments use two primary retrieval sources:

\begin{itemize}
    \item \textbf{arXiv API}. For research-oriented topics, we query the arXiv API using keyword searches ranked either by relevance or by submission date. The API returns a user-defined number of papers along with metadata including the title, arXiv ID, last updated date, direct link, abstract, authors, category, comments, journal reference, and a PDF link. arXiv is highly dynamic, with thousands of papers added monthly, making it an excellent source for cutting-edge developments. However, this dynamism can also introduce variability in quality, depth, and accessibility, as preprints may not yet be peer-reviewed or aligned with educational standards.
    \item \textbf{Biomedical Textbook Corpus} \citep{jin2021disease}. Used for foundational or pedagogical topics, this corpus contains 125,800 text snippets extracted from 18 curated biomedical textbooks. We use BM25 lexical ranking to retrieve the most relevant snippets for a given topic. BM25 scores each snippet based on term frequency (how often a keyword appears), inverse document frequency (how unique that keyword is across the corpus), and a normalization factor for document length. This ensures that retrieved passages are both topically relevant and concise. Textbooks, while less current than arXiv, provide stable, peer-reviewed explanations that can be trusted for conceptual clarity and educational reliability.
\end{itemize}

The contrast between these two sources is intentional. Retrieval from arXiv allows for recency and topic breadth, enabling SlideBot to cover emerging concepts and evolving research frontiers. On the other hand, textbook retrieval brings stability and pedagogical reliability, ensuring foundational topics are explained in a way that supports structured learning. While our experiments select one source or the other for controlled comparisons, a real-world deployment could easily combine them. An alternative retrieval agent could pair different retrievers with varied corpora, enabling customization for specific domains or formats. For example, multimodal retrieval could incorporate figures, diagrams, or videos alongside text, while domain-specific APIs or curated datasets could replace or augment current sources to better match instructional needs.

To illustrate how the Retriever Agent contributes to the final presentation, we present one of the summaries used to assist the creation of SlideBot's presentation in Figure \ref{fig:demo_figure}. In this example, the topic "Manifold Learning" led the Moderator to generate targeted search keywords, which retrieved an arXiv paper on genetic programming for manifold learning \citep{lensen2021genetic}. The Retriever Agent then produces the summary presented in Figure \ref{fig:summary_output}. This summary captures both a high-level overview and a deeper breakdown of the paper’s contributions, methodology, and results. The in-depth structure not only supports conceptual accuracy in the generated slides but also helps the downstream planner decide where and how to incorporate the content.

\subsubsection{Code Generator.}

In our experiments, the Moderator constructs a plan for the presentation using the template in Figure \ref{fig:mod_plan_prompt}. The audience is assumed to be unfamiliar with the topic, so introductory and motivational content is explicitly included before technical details. Only the provided source type (textbook or arXiv) is used, preventing hallucination or external inference. Slide content must follow the structural guide in Figure \ref{fig:structural_guide}. The guide itself embeds CLT/CTML principles, as shown in the annotations, which explicitly label how each slide direction aligns with instructional strategies such as intrinsic load management, signaling, and coherence. 

Once the Moderator finalizes the slide plan, it is passed to the Code Generator Agent, which produces the complete \LaTeX{} Beamer source code using the prompt in Figure \ref{fig:cg_code_prompt}. This step focuses on faithfully implementing the planned content, ensuring that every slide follows the specified structure, all facts are directly grounded in the provided sources, and each claim is accompanied by an appropriate in-text citation. An alternative Code Generator agent could move beyond prompt-based generation by using fine-tuned models trained on high-quality Beamer examples or by integrating template libraries for specific instructional styles. 

Figure \ref{fig:plan_output} shows a generated plan, which begins by orienting slides, progressing through core content and case studies, and concluding with critical perspectives and future directions. This structure, generated by the Moderator, already embeds CLT and CTML principles to manage cognitive load and enhance retention. The Code Generator then operationalizes this plan by producing the complete Beamer presentation, ensuring consistent formatting, visual emphasis, and proper citation for each factual statement.

\begin{figure}
\begin{tcolorbox}[colback=gray!10, colframe=black!90, title=Moderator: Slide Plan Creation Prompt Template]
You are tasked with generating a slide-by-slide plan for a graduate-level LaTeX Beamer presentation on the topic: \{topic\}.

The audience is not expected to be deeply familiar with the topic, so include introductory and motivational slides before discussing technical details.

Use only the provided \{source\_type\} as your information source. Do not infer beyond the given material.

Construct the presentation using the following structural guide (adjust section lengths and reorganize as needed):  
\{structural guide\}

For each slide, output:
\begin{itemize}
    \item A slide title
    \item A brief description of what it will contain
\end{itemize}

Maintain logical flow between slides, referencing earlier material where relevant.

Here is the source material:  
\{context\}
\end{tcolorbox}
\caption{Prompt Template instructing the Moderator to create a slide plan following the structural guide.}
\label{fig:mod_plan_prompt}
\end{figure}

\begin{figure*}
\begin{tcolorbox}[colback=gray!10,colframe=black!70,title=Annotated Structural Guide.]

\begin{enumerate}
    \item \textbf{Title} – Presents the topic, instructor, and session context clearly. 
    
    \item \textbf{Roadmap / Table of Contents} – Outlines the structure of the presentation to help learners anticipate the flow. 
    \textit{Signaling:} Highlight section headings to draw attention. 
    \textit{Intrinsic Load Management:} Provide a preview that organizes upcoming information.

    \item \textbf{Introduction to the Topic} – Uses \textbf{textbook-derived content} when available to establish stable, peer-reviewed definitions and foundational concepts, supplemented by paper introductions if needed. 
    \textit{Coherence:} Limit to essential terminology and ideas. 

    \item \textbf{Why This Topic Matters} – Begins with high-level concepts and incorporates recent developments to demonstrate currency and relevance. 
    \textit{Signaling:} Use bolded key terms and application-oriented language. 

    \item \textbf{Motivating Limitations of Early Approaches} – Establish limitations and case-specific drawbacks if applicable. 
    \textit{Coherence:} Present only limitations directly tied to motivating later content. 
    \textit{Signaling:} Highlight the link between each limitation and the need for advancement.

    \item \textbf{Recent Advancements} – Primarily from \textbf{paper-derived summaries}, maintain a consistent, clear structure for each paper. \textit{Signaling:} Each bullet should follow a consistent schema: Contribution/Metric - Impact.

    \item \textbf{Critical Perspectives} – Provide a holistic, balanced view. 
    \textit{Coherence:} Tie critiques back to material presented earlier. 
    \textit{Signaling:} Use parallel phrasing to make contrasts explicit.

    \item \textbf{Future Directions} – Draws mainly from paper discussion and future work sections, optionally integrating textbook outlooks for long-term trends.  
    \textit{Germane Load:} Encourage synthesis with previously discussed content.

    \item \textbf{Conclusion} – Reinforces 3–5 key takeaways aligned with learning objectives. 
    \textit{Signaling:} Parallel summary bullet structure to strengthen recall. 
    \textit{Coherence:} Avoid redundancy with earlier slides by focusing on synthesis.

    \item \textbf{References} – Lists sources cited in slides, including paper DOIs/arXiv IDs and textbook details. 
\end{enumerate}

\end{tcolorbox}
\caption{Presentation structural guide used to aid the Moderator in constructing a slide plan. Various directions within the guide are labeled with corresponding italicized principles from CLT and CTML. }
\label{fig:structural_guide}
\end{figure*}

\begin{figure}
\begin{tcolorbox}[colback=gray!10, colframe=black!90, title=Code Generator: Slide Creation Prompt Template]

You are tasked with generating a full LaTeX Beamer presentation based on the following slide plan. 

{plan}

Your job is to write high-quality slide code in LaTeX using the Beamer package. You must follow these formatting and content guidelines carefully:

- Each slide (except the title and references) should have a descriptive title and 3–5 bullet points.
- Each bullet point should be concise, informative, and grounded in the source content.
- Use bolding (textbf{{}}), italics (textit{{}}), coloring, and other LaTeX formatting to enhance readability and emphasis when appropriate.
- Add a cite{{}} next to each bullet point that includes a fact or claim, referencing the appropriate source.  

Source Material: 
Use only the following source texts. Make sure every fact presented is supported and cited properly using 'cite{{}}'. 
{context}

Topic: {topic}

\end{tcolorbox}
\caption{Prompt Template for the Code Generator to output the \LaTeX code for the presentation following the plan given by the Moderator.}
\label{fig:cg_code_prompt}
\end{figure}

\begin{figure*}
\begin{tcolorbox}[colback=gray!10,colframe=blue!70,title=Slide Plan Case Study ,fonttitle=\bfseries]

\textbf{Slide 1: Title}

- Description: Introduce the title of the presentation, "Manifold Learning," along with the presenter's name and affiliation.

\textbf{Slide 2: Roadmap}

- Description: Outline the structure of the presentation, highlighting the key sections that will be covered, including an introduction, motivation, recent advancements, and future directions.

\textbf{Slide 3: Introduction to the Topic}

- Description: Define manifold learning as a subfield of machine learning focused on understanding high-dimensional data by identifying its underlying low-dimensional structures.

\textbf{Slide 4: Why This Topic Matters}

- Description: Discuss the significance of manifold learning in handling complex high-dimensional data, particularly in applications like multimedia, where traditional methods may struggle.

\textbf{Slide 5: Motivating Limitations of Early Approaches}

- Description: Highlight the limitations of early manifold learning techniques, such as their inability to preserve local topology and the challenges posed by noise in data.

\textbf{Slide 6: Recent Advancements in Manifold Learning}

- Description: Provide an overview of recent advancements in manifold learning, setting the stage for specific algorithms and methodologies that have emerged.

\textbf{Slide 7: Overview of Isomap}

- Description: Introduce Isomap, explaining its approach to maintaining the intrinsic geometry of data using geodesic distances, and its role in dimensionality reduction.

\textbf{Slide 8: Overview of Locally Linear Embedding (LLE)}

- Description: Describe LLE, focusing on its method of reconstructing data points based on their neighbors and minimizing reconstruction error.

\textbf{Slide 9: Overview of t-Distributed Stochastic Neighbor Embedding (t-SNE)}

- Description: Explain t-SNE, emphasizing its effectiveness in visualizing high-dimensional data in lower dimensions and its use of Kullback-Leibler divergence for optimization.

\textbf{Slide 10: Genetic Programming for Manifold Learning}

- Description: Introduce the concept of using genetic programming to enhance manifold learning, particularly in preserving local topology during the embedding process.

\textbf{Slide 11: Results of Genetic Programming Approach}

- Description: Present the results of the genetic programming approach, comparing its performance against traditional methods like t-SNE and UMAP, highlighting improvements in local topology preservation.    

\textbf{Slide 12: Functorial Manifold Learning}

- Description: Introduce the functorial framework for manifold learning, explaining how algorithms can be treated as functors that map pseudometric spaces to optimization objectives.

\textbf{Slide 13: Stability Analysis in Functorial Learning}

- Description: Discuss the stability analysis of manifold learning algorithms using interleaving distance, providing bounds that quantify the impact of noise on embeddings.

\textbf{Slide 14: New Algorithms from Functorial Framework}

- Description: Highlight the new algorithms developed from the functorial framework, showcasing their competitive performance against existing methods.

\textbf{Slide 15: Critical Perspectives}

- Description: Offer a critical perspective on the advancements in manifold learning, discussing potential drawbacks and areas where further research is needed.

\textbf{Slide 16: Future Directions}

- Description: Explore future directions for manifold learning, including potential improvements in algorithm robustness and applications in emerging fields.

\textbf{Slide 17: Conclusion}

- Description: Summarize the key points discussed in the presentation, reiterating the importance of manifold learning and its advancements.

\textbf{Slide 18: References}

- Description: Provide a list of references used throughout the presentation, ensuring proper citation of the discussed papers and methodologies.

\end{tcolorbox}
\caption{Slide Plan constructed by the Moderator for the presentation "Manifold Learning." }
\label{fig:plan_output}
\end{figure*}

\begin{figure*}
\begin{tcolorbox}[colback=gray!10, colframe=black!90, title=Enhancer: Embedded Comment Prompt Template]

You are an expert enhancing a set of graduate-level slides by adding togglable instructor comments using the `pdfcomment` package in LaTeX.

Instructions:
\begin{itemize}
    \item 1. Use the provided slide code as a base and refer to the additional context to support the comments or slide content. Ensure all information is accurate and cited.
    \item 2. Add togglable comments using `pdfcomment` with these types:
    \begin{itemize}
        \item Conceptual suggestions (e.g., key takeaways, simplified explanations). 
        \item Teaching strategies (e.g., questions, real-world applications).
        \item Student context suggestions (e.g., misconceptions, level adjustments).
        \item Lecture flow suggestions (e.g., transitions, time management).
        \item Textbook integration (e.g., related textbook sections or figures).
        \item Discussion/activity ideas (e.g., case studies, group discussions).
        \end{itemize}

    \item 3. Frame comments as optional suggestions to help instructors customize the slides collaboratively.
    \item 4. Ensure \textbackslash cite commands on slides align with properly formatted references on the last slide.
    \item 5. Always place comments within the itemize container and before any figure macros. 
    \item 6. Do not alter the preamble.
\end{itemize}
Inputs:
\begin{itemize}
    \item Slide Code: {slidecode}
    \item Additional Context: {context}
    \item Topic: {topic}
\end{itemize}

Output:
Provide LaTeX code for the slides with `pdfcomment` comments added, ensuring comments are helpful, concise, and togglable.

\end{tcolorbox}
\caption{Prompt Template guiding the Enhancer to insert appropriate figures into the presentation.}
\label{fig:enhancer_comment_prompt}
\end{figure*}

\begin{figure*}
\begin{tcolorbox}[colback=gray!10, colframe=black!90, title=Enhancer: Figure Prompt Template]

You are an expert in scientific presentation design. Your task is to improve an existing LaTeX Beamer slide deck by inserting figures and optionally formulas or pseudocode using custom macros.

Available Macros (use only when appropriate):

\begin{itemize}
    \item 1. \textbackslash drawpipeline\{Number of Steps\}\{Step1, Step2, Step3, ...\}  
            — Use for sequential processes or architecture pipelines.
    \item 2. \textbackslash inlineformula\{ \textbackslash [ formula \textbackslash ] \} - Use to display important mathematical expressions (e.g., update rules, error terms). If the code already has \[\] or similar, replace them with this macro.
    \item 3. \textbackslash inlinepseudocode\{ ... \} - Use to illustrate algorithms or training steps in pseudocode (via `algorithm2e`).
    \item 4. \textbackslash drawconfmat\{TN\}\{FP\}\{FN\}\{TP\} - Use for confusion matrix or classifier performance.
    \item 5. \textbackslash drawnetwork\{layer1, layer2, ..., layerN\} - Use to visualize a feedforward neural network with the given number of neurons per layer  (e.g., \textbackslash drawnetwork\{2,3,4,1\} draws a 4-layer network).
    \item 6. \textbackslash drawgenericplot\{xlabel\}\{ylabel\}\{function1\}\{function2\}\{function3\}\{legend\} - Use only when you want to show actual numeric trends or comparisons (e.g., bias vs variance curves).
\end{itemize}

Instructions:

\begin{itemize}
    \item 1. Carefully read each slide's content and the source context to identify if any of the above macros would meaningfully enhance understanding.
    \item 2. You may insert at most one figure macro per slide, placed after bullet points and before `\textbackslash end\{frame\}`. Not all slides need a figure.
    \item 3. You may also insert at most one inline formula or pseudocode macro per slide when relevant, but not if there is already a figure. These may appear inline under a bullet or as a standalone block.
    \item 4. Use meaningful illustrative placeholders based on the topic or context. You don’t need to use real data, but the figure/formula/pseudocode should conceptually match.
    \item 5. Do not alter the rest of the LaTeX structure, such as slide titles, text, existing bullets, or the preamble. Only add macros.
    \item 6. If a slide has no meaningful visual or symbolic enhancement, leave it unchanged.
\end{itemize}

Inputs:
\begin{itemize}
    \item Slide Code: LaTeX Beamer code for slides (including bullet items and citations).
    \item Additional Context: Source material, such as summaries or excerpts from research papers.
    \item Topic: {topic}
\end{itemize}

Output:
Return the entire updated LaTeX code with macro insertions where appropriate. 
"""

\end{tcolorbox}
\caption{Prompt Template instructing the Enhancer to embed instructor comments into the presentation.}
\label{fig:enhancer_figure_prompt}
\end{figure*}

\begin{figure*}[ht]
\centering
\includegraphics[width=\linewidth]{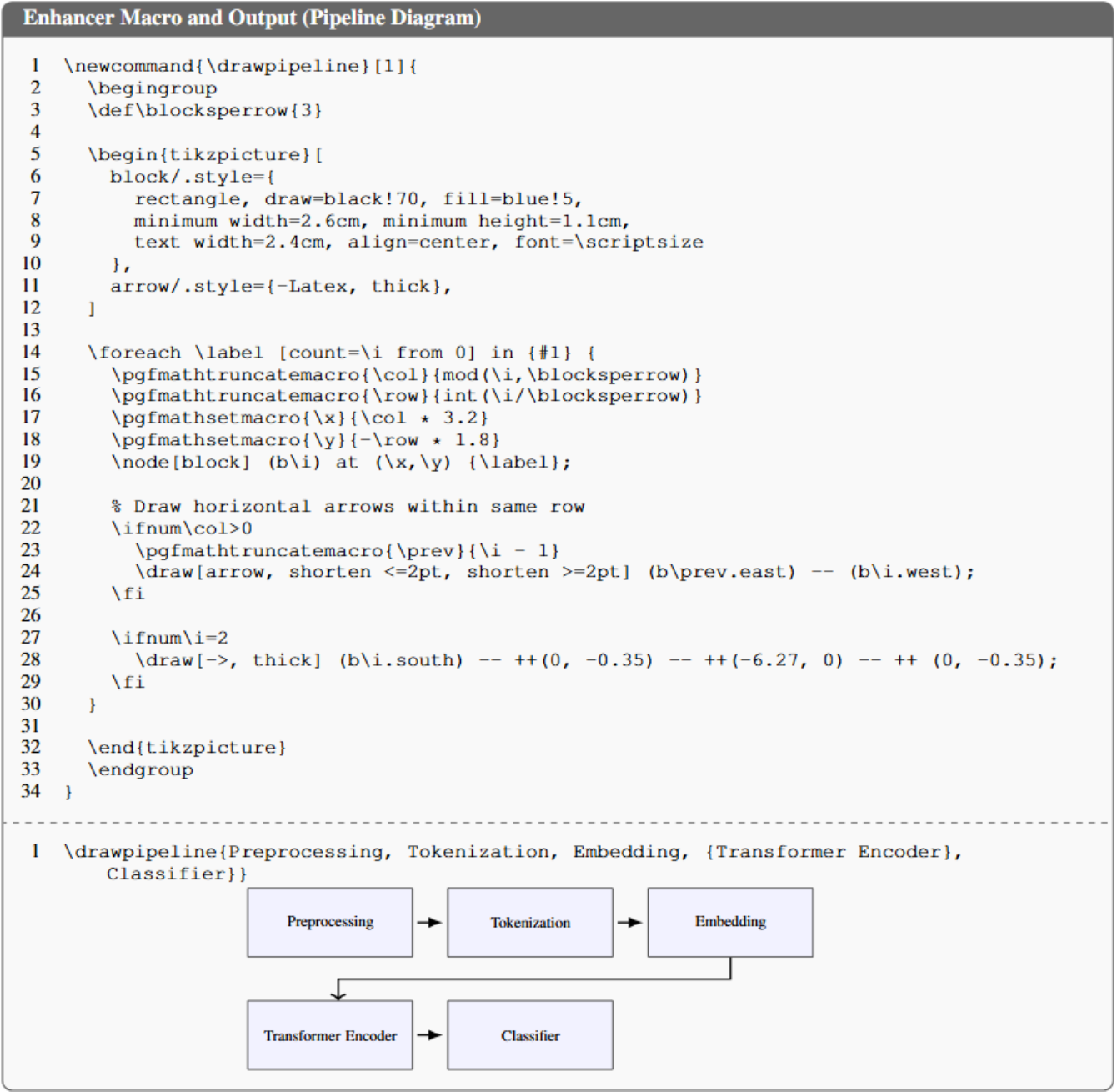}
\caption{The Enhancer’s prewritten line graph macro (top) and the rendered output generated by the corresponding code (bottom).}
\label{fig:pipeline_macro}
\end{figure*}

\begin{figure*}[ht]
\centering
\includegraphics[width=\linewidth]{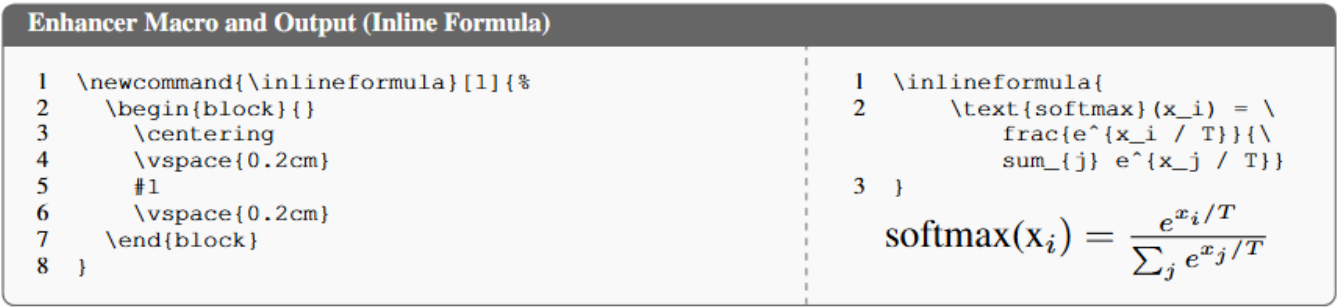}
\caption{Side-by-side view of the Enhancer’s prewritten inline formula macro (left) and the rendered output generated by the corresponding code (right).}
\label{fig:formula_macro}
\end{figure*}

\begin{figure*}[ht]
\centering
\includegraphics[width=\linewidth]{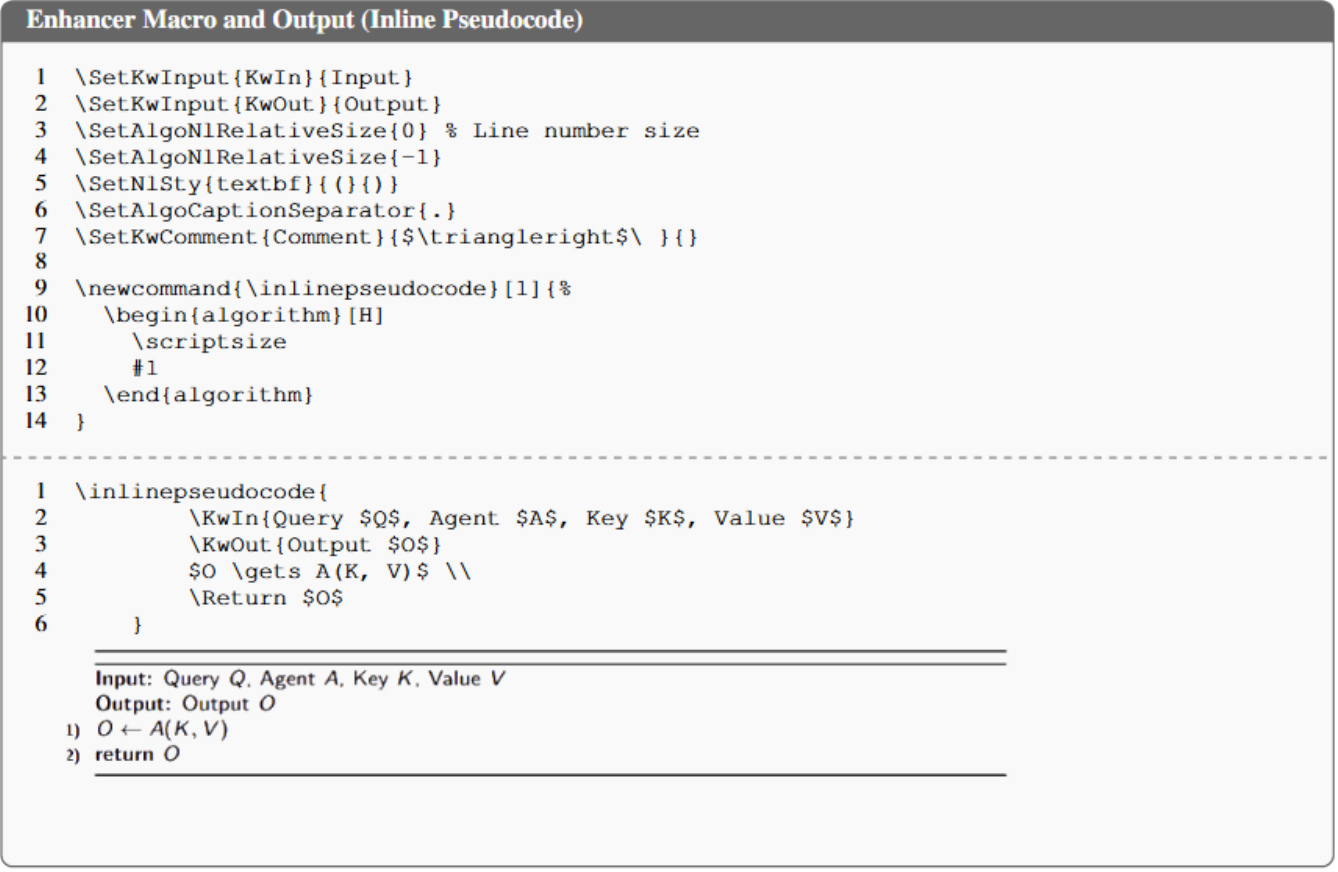}
\caption{The Enhancer’s prewritten inline pseudocode macro (top) and the rendered output generated by the corresponding code (bottom).}
\label{fig:pseudocode_macro}
\end{figure*}

\begin{figure*}[ht]
\centering
\includegraphics[width=\linewidth]{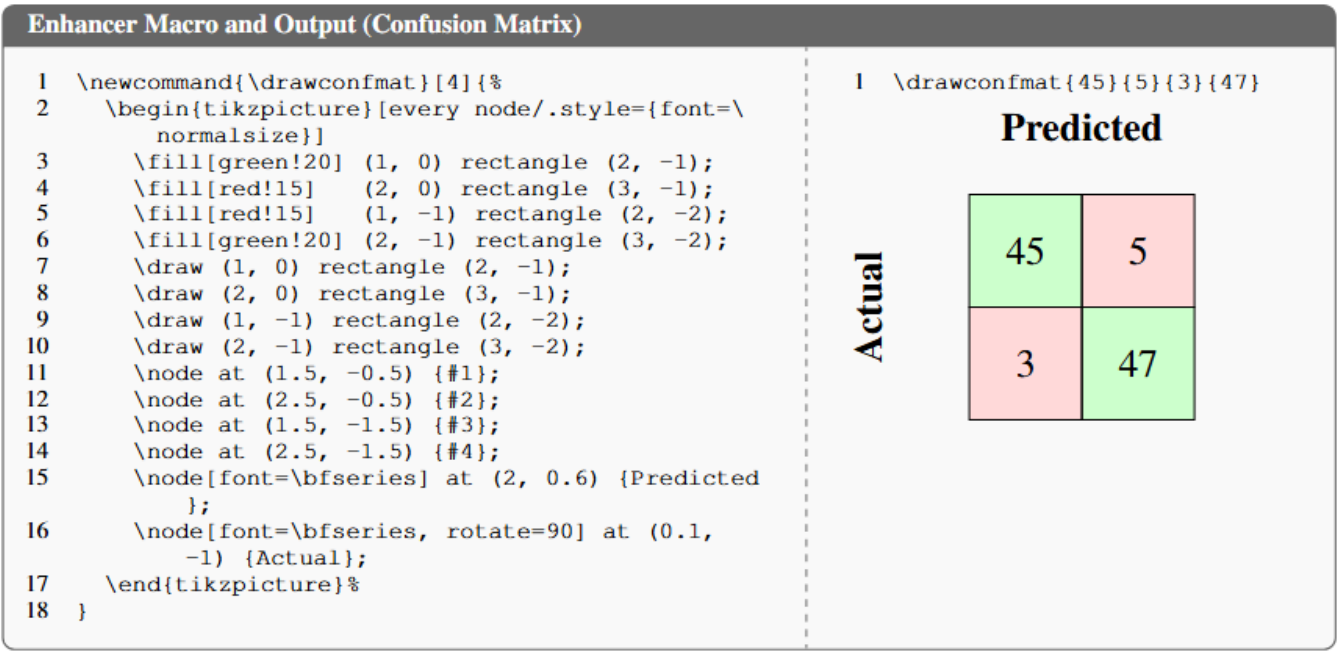}
\caption{Side-by-side view of the Enhancer’s prewritten confusion matrix macro (left) and the rendered output generated by the corresponding code (right).}
\label{fig:conf_macro}
\end{figure*}

\begin{figure*}[ht]
\centering
\includegraphics[width=\linewidth]{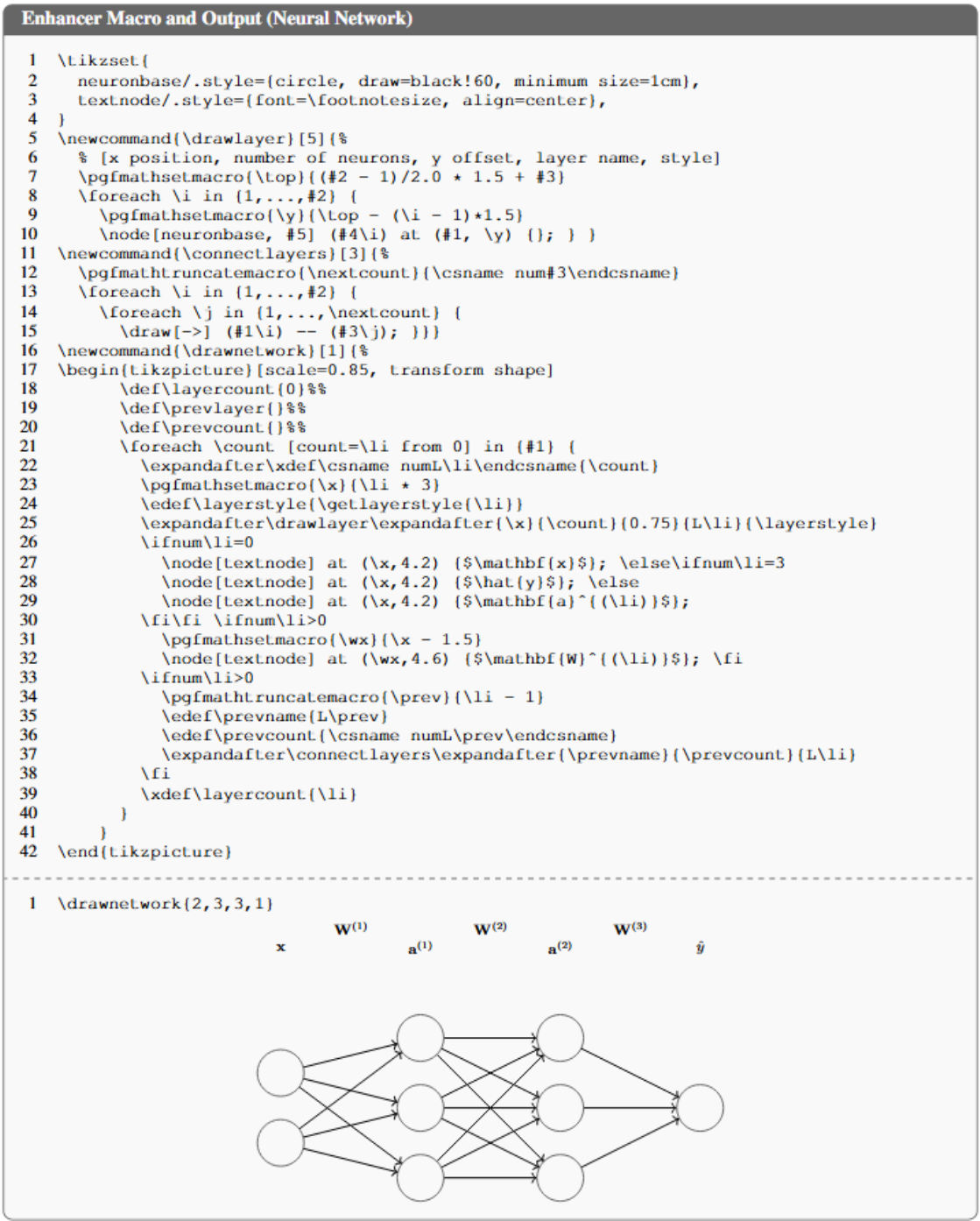}
\caption{Enhancer’s prewritten neural network macro (top) and the rendered output generated by the corresponding code (bottom).}
\label{fig:network_macro}
\end{figure*}

\begin{figure*}[ht]
\centering
\includegraphics[width=\linewidth]{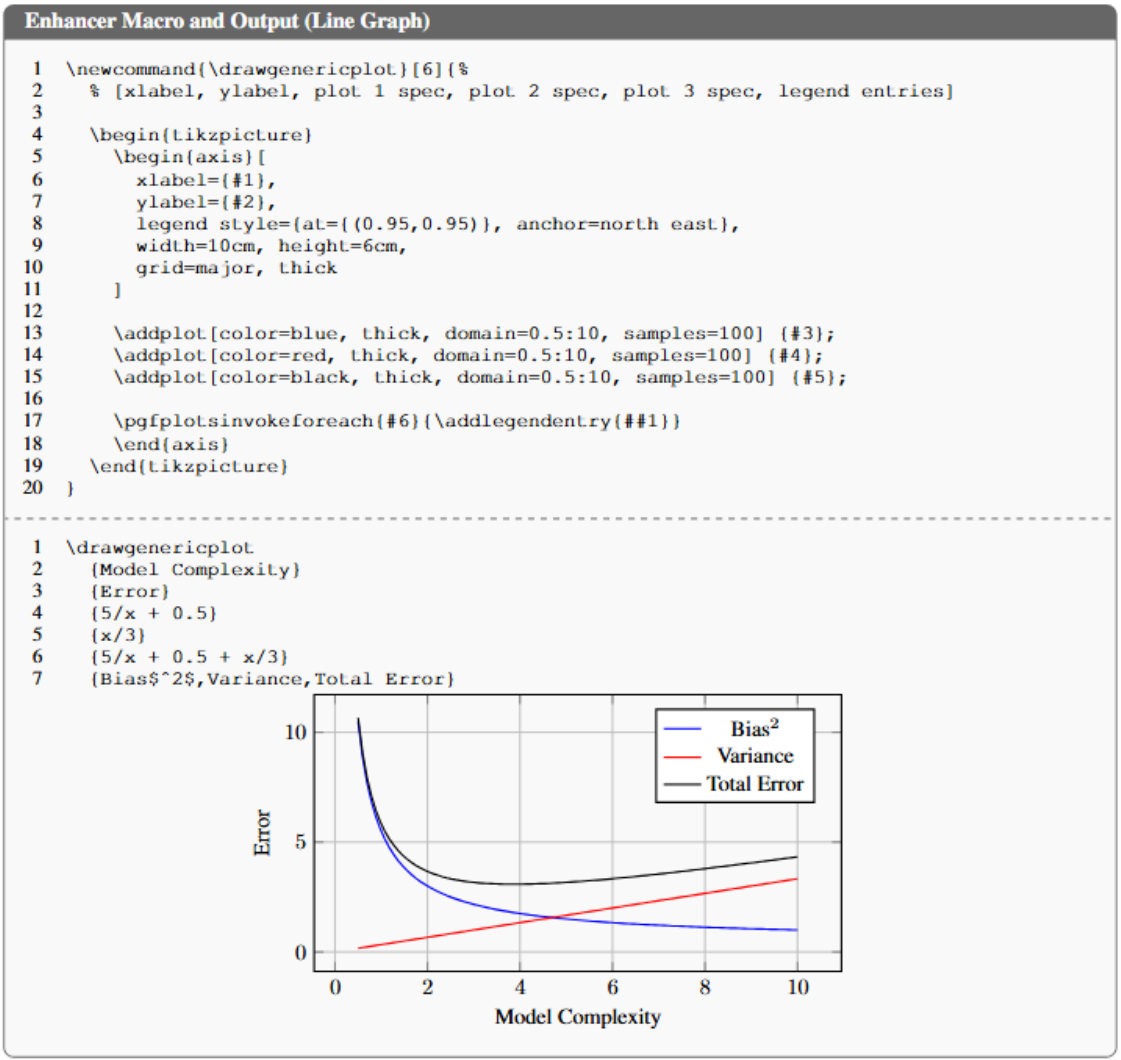}
\caption{The Enhancer’s prewritten line graph macro (top) and the rendered output generated by the corresponding code (bottom).}
\label{fig:curve_macro}
\end{figure*}

\subsubsection{Enhancer.}

The Enhancer agent improves the pedagogical and visual quality of AI-generated slide decks by inserting instructor-oriented comments and meaningful visual elements. It operates under two specialized prompt templates: one for embedded comments (Figure \ref{fig:enhancer_comment_prompt}) and another for figures/macros (Figure \ref{fig:enhancer_figure_prompt}), each targeting a different layer of instructional support. The comment enhancement process is designed to support instructors with optional, togglable annotations via the pdfcomment package.

The code for the figure macros and example implementations are shown in Figures \ref{fig:pipeline_macro}-\ref{fig:curve_macro}. These macros serve two primary purposes. First, as illustrated in the figures, the underlying LaTeX required to produce each visual is relatively long and complex, whereas the macro call itself is short and concise. This greatly simplifies the generation process for the model, reducing the likelihood of syntax errors and enabling it to focus on selecting an appropriate visual rather than constructing the figure code from scratch. Second, for elements such as inline formulas and pseudocode, the model can readily produce the raw content, but defining macros enforces a consistent output format and style. This uniformity is especially important for maintaining cognitive load consistency and alignment, ensuring that all generated visuals integrate smoothly into the slide deck’s overall design. The final "Manifold Learning" presentation is displayed in the following section.

\subsection{Supplementary Presentation Examples}

We present full presentations generated by SlideBot, Copilot, and the zero-shot method. Comments are not displayed due to system settings. Presentations displayed are
\begin{itemize}
    \item "Manifold Learning" by SlideBot
    \item "Multi-Head Attention" by SlideBot
    \item "Multi-Head Attention" by Copilot
    \item "Neurotransmitters" by Zero-shot Generation
\end{itemize}

\FloatBarrier

\includepdf[
  pages=1-6,
  nup=2x3,
  frame,
  scale=0.92,
  pagecommand={
      \begin{tikzpicture}[remember picture,overlay]
        \def\cellw{.5\paperwidth} 
        \def\cellh{.333333\paperheight}
        \def\xin{6mm} \def\yin{5mm}
        \node[anchor=south east] at (current page.north west)
          [xshift=\cellw-\xin-30mm, yshift=-\cellh+\yin+60mm]{\large\textbf{{``Manifold Learning" by SlideBot}}};
      \end{tikzpicture}
  }
]{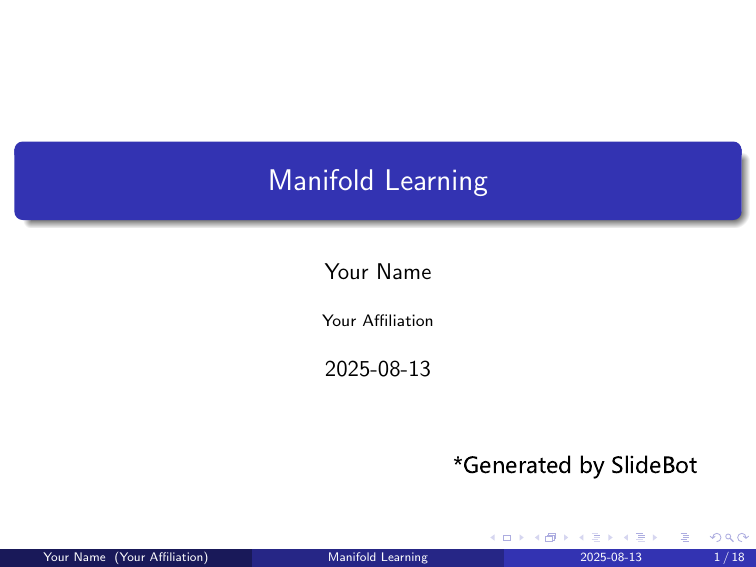}

\includepdf[
  pages=7-, nup=2x3, frame, scale=0.92,
  pagecommand={}
]{appendix_figs/example_presentations/SlideBot_manifold.pdf}

\includepdf[
  pages=1-6,
  nup=2x3,
  frame,
  scale=0.92,
  pagecommand={
      \begin{tikzpicture}[remember picture,overlay]
        \def\cellw{.5\paperwidth} 
        \def\cellh{.333333\paperheight}
        \def\xin{6mm} \def\yin{5mm}
        \node[anchor=south east] at (current page.north west)
          [xshift=\cellw-\xin-30mm, yshift=-\cellh+\yin+60mm]
          {\large\textbf{``Multi-Head Attention" by SlideBot}};
      \end{tikzpicture}
  }
]{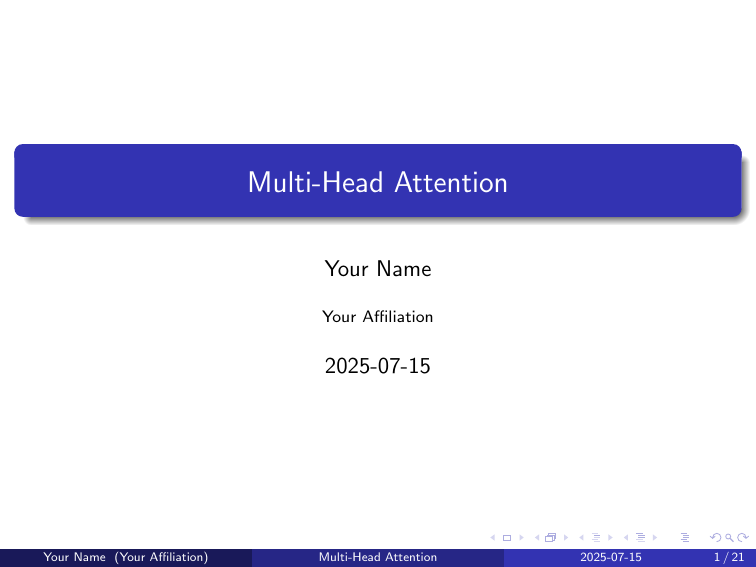}

\includepdf[
  pages=7-, nup=2x3, frame, scale=0.92,
  pagecommand={}
]{appendix_figs/example_presentations/SlideBot_multi-head.pdf}

\includepdf[
  pages=1-6,
  nup=2x3,
  frame,
  scale=0.92,
  pagecommand={
      \begin{tikzpicture}[remember picture,overlay]
        \def\cellw{.5\paperwidth} 
        \def\cellh{.333333\paperheight}
        \def\xin{6mm} \def\yin{5mm}
        \node[anchor=south east] at (current page.north west)
          [xshift=\cellw-\xin-30mm, yshift=-\cellh+\yin+33mm]
          {\large\textbf{''Multi-Head Attention" by Copilot}};
      \end{tikzpicture}
  }
]{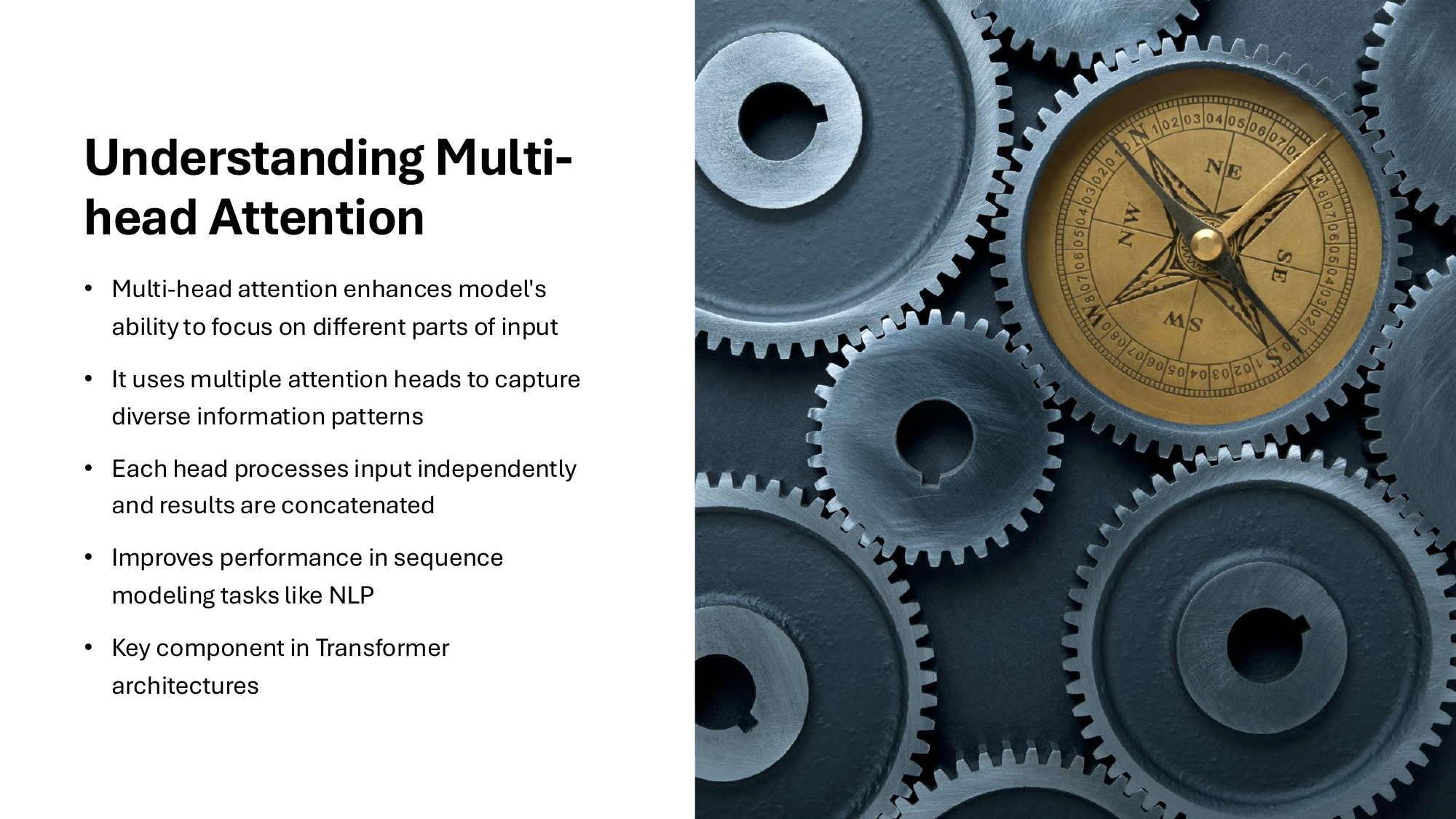}

\includepdf[
  pages=7-, nup=2x3, frame, scale=0.92,
  pagecommand={}
]{appendix_figs/example_presentations/Copilot_multi-head.pdf}

\includepdf[
  pages=1-6,
  nup=2x3,
  frame,
  scale=0.92,
  pagecommand={
      \begin{tikzpicture}[remember picture,overlay]
        \def\cellw{.5\paperwidth} 
        \def\cellh{.333333\paperheight}
        \def\xin{6mm} \def\yin{5mm}
        \node[anchor=south east] at (current page.north west)
          [xshift=\cellw-\xin-15mm, yshift=-\cellh+\yin+60mm]
          {\large\textbf{''Neurotransmitters" using Direct Prompting}};
      \end{tikzpicture}
  }
]{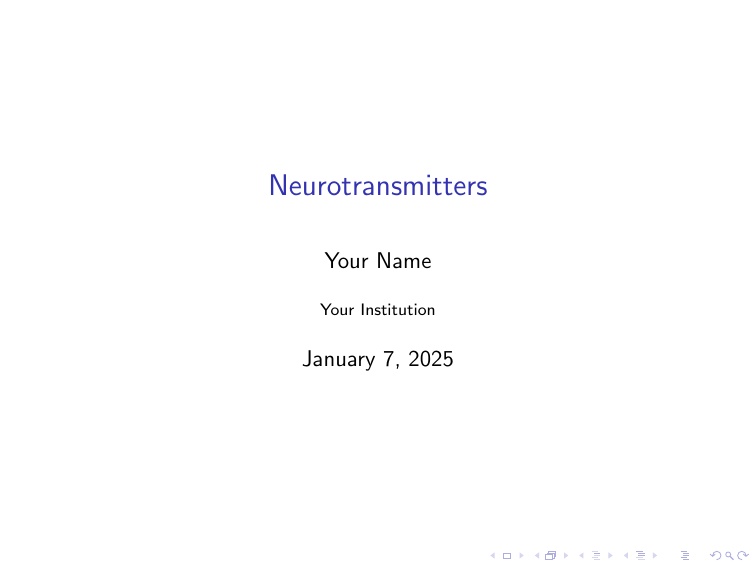}

\includepdf[
  pages=7-, nup=2x3, frame, scale=0.92,
  pagecommand={}
]{appendix_figs/example_presentations/Zero_neurotransmitters.pdf}

\end{document}